\documentclass[orivec]{llncs}
\usepackage{IEEEtrantools}

\usepackage{mathrsfs}
\usepackage{amssymb}
\usepackage[cmex10]{amsmath}
\usepackage{amsthm}
\usepackage{graphics}
\usepackage{latexsym}
\usepackage{amsfonts}
\usepackage{color}
\usepackage{ctable}
\usepackage{graphicx}
\usepackage{verbatim}
\usepackage{epstopdf}
\usepackage{dsfont}

\newtheorem{theo}{Theorem}

\newtheorem{coro}[theo]{Corollary}

\theoremstyle{remark}
\newtheorem{exam}{Example}
\newtheorem{rema}[theo]{Remark}

\newcommand{\be}{\begin{IEEEeqnarray*}{rCl}}
\newcommand{\ee}{\end{IEEEeqnarray*}}
\newcommand{\ben}{\begin{IEEEeqnarray}{rCl}}
\newcommand{\een}{\end{IEEEeqnarray}}
\newcommand{\lb}[1]{\left[\begin{array}{#1}}
\newcommand{\rb}{\end{array}\right]}
\newcommand{\lp}[1]{\left(\begin{array}{#1}}
\newcommand{\rp}{\end{array}\right)}

\newcommand{\leftd}[1]{\left\{\begin{array}{#1}}
\newcommand{\rightd}{\end{array}\right.}

\def\A {\mathbf{A}}
\def\B {\mathbf{B}}
\def\C {\mathbf{C}}

\def\G {\mathbf{G}}
\def\H {\mathbf{H}}
\def\I {\mathbf{I}}

\def\Q {\mathbf{Q}}
\def\R {\mathbf{R}}

\def\W {\mathbf{W}}

\def\a {\mathbf{a}}
\def\b {\mathbf{b}}
\def\cc {\mathbf{c}}

\def\s {\mathbf{s}}

\def\w {\mathbf{w}}
\def\x {\mathbf{x}}
\def\y {\mathbf{y}}

\def\Ds {\mathscr{D}}

\def\Nc {\mathcal{N}}

\def\Sc {\mathcal{S}}

\def\PSI {\boldsymbol{\psi}}

\def\muu{\boldsymbol{\mu}}
\def\teta{\boldsymbol{\theta}}

\def\Rb {\mathbb{R}}

\def\Eb {\mathbb{E}}

\newcommand{\sign}{\mathrm{sign}}

\newcommand{\CD}{\xrightarrow[N\rightarrow\infty]{\Ds}}

\newcommand{\defeq}{\stackrel{\mathrm{def}}{=}}
\newcommand{\TR}{\mathsf{T}}
\newcommand{\argmin}[1]{\underset{#1}{\operatorname{argmin\,\,}}}

\newcommand{\cov}{\mathrm{Cov}}

\newcommand{\EE}{\widehat{\Eb}}

\title{An Overview of the Asymptotic Performance of the Family of the FastICA Algorithms}
\author{Tianwen~Wei}
\institute{Laboratoire de Math\'ematiques de Besancon \\ Universit\'e de Franche-Comt\'e \\
 16 Route de Gray \\
 25000 Besancon, France}

\begin{document}
\maketitle

\begin{abstract}
This contribution summarizes the results on the asymptotic performance of several variants of the FastICA algorithm. A number of new closed-form expressions 
are presented. 
\end{abstract}

\begin{keywords}
Independent component analysis, symmetric FastICA, deflationary FastICA, data whitening, data centering, asymptotic performance.
\end{keywords}

\section{Introduction}

In what follows, we denote scalars by lowercase letters $(a,b,c,\ldots)$, vectors by
boldface lowercase letters $(\a,\b,\cc, \ldots)$ and matrices by
 boldface uppercase letters $(\A,\B,\C,\ldots)$. Greek letters $(\alpha,\beta,\gamma,\ldots)$ are reserved for particular scalar quantities. We denote by $\A^\TR$ the matrix transpose of $\A$ and by $\|\cdot\|$ the Euclidean norm.

\subsection{ICA Data Model\label{sectionIIA}}
We consider the following noiseless linear ICA model:
\be
\y(t)=\H\s(t),\quad t=1,\ldots,N,\label{ICAmodel1}
\ee
where
\begin{enumerate}
\item $\s(t)\defeq(s_1(t),\ldots, s_d(t))^\TR$ denotes the $t$th realization of the unknown \emph{source signal}. The components $s_1(t),\ldots, s_d(t)$ are mutually statistically independent, have unit variance and at most one of them is Gaussian. 
Furthermore, $\s(1),\ldots, \s(N)$ denote $N$ independent realizations of $\s$.
\item $\y(t)\defeq(y_1(t),\ldots, y_d(t))^\TR$ denotes the $t$th realization of the \emph{observed signal}.
\item $\H\in\Rb^{d\times d}$ is a full rank square matrix, called the mixing matrix.
\end{enumerate}

\subsection{Data Preprocessing}
Most ICA methods require the observed signal $\{\y(t)\}$ to be standardized 
\cite{COMO94,CARD1993,HYVA99}. The standardization of $\{\y(t)\}$ consists of the data centering and data whitening, which involve the estimation of $\Eb[\y]$ and $\cov(\y)$. 
In practice, $\Eb[\y]$ and $\cov(\y)$ are usually estimated by the sample mean and sample variance:
\be
\bar{\y}\defeq \sum_{t=1}^N \frac{1}{N}\y(t), \quad
\widehat{\C} \defeq \frac{1}{N}\sum_{t=1}^N (\y(t) - \bar{\y})(\y(t) - \bar{\y})^\TR.
\ee  
 In this work, we shall consider several different data preprocessing scenarios. 
 Denote
 \be
 \widetilde{\C}= \frac{1}{N}\sum_{t=1}^N (\y(t) - \Eb[\y])(\y(t) - \Eb[\y])^\TR.
 \ee
The following data preprocessing scenarios will be studied: 
\begin{enumerate}
\item[1).] Theoretical whitening and theoretical centering.
 \ben\label{Preprocess1}
 \x(t)  \defeq \cov(\y)^{-\frac{1}{2}}(\y(t)-\Eb[\y]).
 \een
\item[2).] Theoretical whitening and empirical centering.
 \ben\label{Preprocess2}
 \x(t)  \defeq \cov(\y)^{-\frac{1}{2}}(\y(t)-\bar{\y}).
 \een
\item[3).] Empirical whitening and theoretical centering.
 \ben\label{Preprocess3}
 \x(t)  \defeq \widetilde{\C}^{-\frac{1}{2}}(\y(t)-\Eb[\y]).
 \een
\item[4).] Empirical whitening and empirical centering.
 \ben\label{Preprocess4}
 \x(t)  \defeq \widehat{\C}^{-\frac{1}{2}}(\y(t)-\bar{\y}).
 \een
\end{enumerate}
In the sequel, $\x(t)$ will always stand for the standardized signal under one of the  scenarios defined above. The specific data preprocessing scenario will be stated explicitly when necessary. 

\subsection{Variants of the FastICA Algorithm}
Before proceeding further, we need to introduce some notations first. We denote by $\Sc$ the unit sphere in $\Rb^d$. We denote by $g(\cdot): \Rb\to\Rb$ the nonlinearity function, and by
$G(\cdot)$ its primitive. The nonlinearity function $g$ is usually supposed to be
 non-linear, non-quadratic and smooth.
For any function $f:\Rb^d\to\Rb^m$, we write
$\EE_{\x}[f(\x)]\defeq \frac{1}{N} \sum_{t=1}^N f(\x(t))$ for conciseness.

\subsubsection{The Deflationary FastICA Algorithm}
This version of the FastICA algorithm extracts the sources sequentially. It consists of the following steps \cite{HYVA99}:
\begin{enumerate}
\item[-] {\bf Input:} $\x(1),\ldots,\x(N)$.
\item[1).] Set $p=1$.
\item[2).] Choose  an arbitrary  initial iterate $\w\in\Sc$;
\item[3).] Run iteration
\ben
\w & \leftarrow & \EE_{\x}[g'(\w^{\TR}\x)\w - g(\w^{\TR}\x)\x] \label{52} \\
\w &\leftarrow &\w-\sum_{i=1}^{p-1}(\w_i^{DFL})^\TR\w \label{52c} \\
\w & \leftarrow & \frac{\w}{\|\w\|} \label{52a}
\een
 until convergence\footnote{We impose the number of iterations to be even, so that the well known sign-flipping phenomenon disappears.  }. The limit is stored as $\w^{DFL}_p$.
 \item[4)] Break if $p=d$. Otherwise $p \leftarrow p+1$
  then go to step 2).
 \item[-] {\bf Output:} $\W^{DFL}=(\w_1^{DFL},\ldots,\w_d^{DFL})$.
\end{enumerate}

\subsubsection{The Symmetric FastICA Algorithm}
The symmetric version of FastICA extracts all the sources simultaneously. 
It can be described as follows:
\begin{enumerate}
\item[-] {\bf Input:} $\x(1),\ldots,\x(N)$.
\item[1).] Choose an arbitrary orthonormal matrix $\W=(\w_1,\ldots,\w_d)\in\Rb^{d\times d}$.
\item[2).] Run
\ben
\w_1 & \leftarrow & \EE_\x[g'(\w_1^{\TR}\x)\w_1 - g(\w_1^{\TR}\x)\x]  \label{127c1} \\
 & \vdots &  \nonumber \\
\w_d & \leftarrow & \EE_\x[g'(\w_d^{\TR}\x)\w_1 - g(\w_1^{\TR}\x)\x]  \label{127c2} \\
\W   & \leftarrow  &    \Big(\W\W^{\TR}  \Big)^{-1/2}\W\label{127b}
\een
 until convergence. The limit is denoted by $\W^{SYM}$.
\item[-] {\bf Output:} $\W^{SYM}=(\w_1^{SYM},\ldots,\w_d^{SYM})$.
\end{enumerate}

\section{Asymptotic Performance}
Let us introduce the notion of gain matrix:
\be
\G^{DFL}\defeq (\W^{DFL})^\TR\C^{-1/2}\H, \quad \G^{SYM}\defeq (\W^{SYM})^\TR\C^{-1/2}\H,
\ee
where $\C^{-1/2}$ stands for the sphering matrix used in the data preprocessing stage,
i.e. $\C=\cov(\y)$ in scenarios (\ref{Preprocess1}) and (\ref{Preprocess2}),
$\C=\widetilde{\C}$ 
in scenario (\ref{Preprocess3})
and  $\C=\widehat{\C}$ in scenario (\ref{Preprocess4}).
Without loss of generality, we shall omit the permutation and sign ambiguities of ICA.
 Then,
%
%
 $\G^{DFL} \approx \I$ and $\G^{SYM} \approx \I$, hence $\C^{-1/2}\W^{DFL}$
 and $\C^{-1/2}\W^{SYM}$ can be considered as estimators of $\B\defeq(\H^{-1})^\TR$.
In the sequel, we will study the asymptotic errors of
$N^{1/2}(\C^{-1/2}\W^{DFL} - \B)$ and $N^{1/2}(\C^{-1/2}\W^{SYM} - \B)$ under proposed data preprocessing scenarios.

The proofs of the results presented below are based on the method of M-estimators.
However, all proofs will be omitted due to the lack of space.  
A complete version of this work can be provided upon request.
 The readers are also referred to \cite{WEI4} for a more detailed account of this subject. 
\subsection{The Asymptotic Error of Deflationary FastICA}
Assume that the  following mathematical expectations exist for $i=1,\ldots,d$:
\ben
\alpha_{i}&\defeq& \Eb[g'(z_{i}) - g(z_{i})z_{i}] \nonumber \\
\beta_{i} & \defeq & \Eb[g(z_{i})^2]  \nonumber  \\
\gamma_{i} & \defeq & \Eb[g(z_{i})z_{i}]  \nonumber \\
\eta_{i} & \defeq & \Eb[g(z_{i})] \nonumber \\
\tau_{i} & \defeq & (\Eb[z_{i}^4]-1)/4, \nonumber
\een
where $z_i=s_i-\Eb[s_i]$ for $i=1,\ldots,d$.
\begin{theo}\label{main1}
Let $\b_i$ denote the $i$th column of $\B$. Under some mild regularity conditions, we have
\be
N^{1/2}(\C^{-1/2}\w_i^{DFL} - \b_i)\CD \Nc(0, \R_{(k)}^{DFL}),
\ee
where $k\in\{1,2,3,4\}$ is the label of the underlying data preprocessing scenario (see (\ref{Preprocess1})-(\ref{Preprocess4})) and $\R_{(k)}^{DFL}$ is given as follows:
\ben
\R_{(1)}^{DFL}& =& \sum_{j=1}^{i-1} \frac{\beta_j^2}{\alpha_j^2}\b_j\b_j^\TR
+ \sum_{ p,q=1\atop p\neq q}^{i-1}\frac{\eta_p\eta_q}{\alpha_p\alpha_{q}}\b_p\b_q^\TR
+   \frac{\beta_i^2}{\alpha_i^2}\sum_{j=i+1}^{d} \b_j\b_j^\TR, \label{DFL1} \\
\R_{(2)}^{DFL}& =&  \sum_{j=1}^{i-1}\frac{\beta_j-\eta_j^2}{\alpha_j^2 }\b_j\b_j^\TR +\frac{\beta_i-\eta_i^2}{\alpha_i^2 } \sum_{j=i+1}^d\b_j\b_j^\TR, \label{DFL2}\\
\R_{(3)}^{DFL}& =&  \sum_{j=1}^{i-1}\frac{\beta_j - \gamma_j^2  + \alpha_j^2}{\alpha_j^{2}} \b_j\b_j^\TR
 +  \sum_{p,q=1\atop p\neq q}^{i-1} \frac{\eta_p\eta_q}{ \alpha_p\alpha_q}\b_p\b_q^\TR  
 \nonumber   + \tau_i  \b_i\b_i^\TR  \\
 && +\frac{\beta_i - \gamma_i^2}{\alpha^2_{i}}\sum_{j=i+1}^{d} \b_j\b_j^\TR   -  \sum_{j=1}^{i-1}\frac{\Eb[s_i^3]\eta_j}{\alpha_j}(\b_j{\b_i^\TR} + \b_i{\b_j^\TR} ) \label{DFL3}, \\
\R_{(4)}^{DFL}&=& \sum_{j=1}^{i-1}\frac{\beta_j - \gamma_j^2  + \alpha_j^2 - \eta_j^2}{\alpha_j^{2}} \b_j\b_j^\TR   + \tau_i \b_i\b_i^\TR 
  +\frac{\beta_i - \gamma_i^2 -\eta_i^2}{\alpha^2_{i}}\sum_{j=i+1}^{d} \b_j\b_j^\TR 
  \nonumber \\
&& -  \sum_{j=1}^{i-1}\frac{\Eb[s_i^3]\eta_j}{\alpha_j}(\b_j{\b_i^\TR} + \b_i{\b_j^\TR} ).\label{DFL4}
\een
\end{theo}

\begin{coro}\label{coroDFL2}There holds
$
N^{1/2}({\G}_{ij}^{DFL} - \delta_{ij})\CD \Nc(0, V_{(k)}^{DFL}),
$
where ${\G}_{ij}^{DFL}$ denotes the $(i,j)$th entry of $\G^{DFL}$ and $V_{(k)}^{DFL}$ is given as follows:
\begin{enumerate}
\item Case $j< i$:
\be
V^{DFL}_{(1)} &=& \frac{\beta_j^2}{\alpha_j^2}   \\
V^{DFL}_{(2)} &=& \frac{\beta_j - \eta_j^2}{\alpha_j^2} \\
V^{DFL}_{(3)} &=& \frac{\beta_j - \gamma_j^2 + \alpha_j^2}{\alpha_j^2} \\
V^{DFL}_{(4)} &=& \frac{\beta_j - \gamma_j^2 + \alpha_j^2 - \eta_j^2}{\alpha_j^2}.
\ee
\item Case $j= i$:
\be
V^{DFL}_{(1)} =V^{DFL}_{(2)} =0 ,\quad\quad V^{DFL}_{(3)} =V^{DFL}_{(4)} = \tau_i .
\ee
\item  Case $j> i$:
\ben
V^{DFL}_{(1)} &=& \frac{\beta_i}{\alpha_i^2}  \label{dflgain1} \\
V^{DFL}_{(2)} &=& \frac{\beta_i - \eta_i^2}{\alpha_i^2}  \label{dflgain2} \\
V^{DFL}_{(3)} &=& \frac{\beta_i - \gamma_i^2}{\alpha_i^2}  \label{dflgain3} \\
V^{DFL}_{(4)} &=& \frac{\beta_i - \gamma_i^2 - \eta_i^2}{\alpha_i^2}. \label{dflgain4}
\een
\end{enumerate}
\end{coro}

\subsection{The Asymptotic Error of Symmetric FastICA}
\begin{theo}\label{main2}
Under some mild regularity conditions, we have
$
N^{1/2}(\C^{-1/2}\w_i^{SYM} - \b_i)\CD \Nc(0, \R_{(k)}^{SYM}),
$
where
\ben
\R^{SYM}_{(1)}&=&\sum_{j\neq i}^d \frac{\beta_i + \beta_j -2\gamma_i\gamma_j -2 \eta_j^2}{(|	\alpha_i|+|\alpha_j|)^2}\b_j\b_j^\TR  + 2\sum_{j\neq i}^d \frac{\eta_j\b_j}{|\alpha_i| + |\alpha_j|} \sum_{j\neq i}^d 	
	\frac{\eta_j\b_j^\TR}{|\alpha_i| + |\alpha_j|}, \label{SYM1}\\
\R^{SYM}_{(2)}&=& \sum_{j\neq i}^d \frac{\beta_i  + \beta_j - 2\gamma_i\gamma_j-2\eta_i^2 }
	{(|\alpha_i| 		+	|\alpha_j|)^2} \b_j\b_j^\TR , \label{SYM2} \\
\R^{SYM}_{(3)} &=&\sum_{j\neq i}^d \frac{\beta_i -\gamma_i^2 + \beta_j -\gamma_j^2 + 	
	\alpha_j^2 - \eta_j^2}{(|\alpha_i| 
 	+ |\alpha_j|)^2} \b_j\b_j^\TR  + \sum_{j\neq i}^d \frac{\eta_j \b_j }{(|\alpha_{i}|+|\alpha_{j})|} \sum_{j\neq i}^d 	\frac{ \eta_j\b_j^\TR }{(|\alpha_{i}|+|\alpha_{j}|)}
	\nonumber \\
&&+ \tau_i\b_i\b_i^\TR 	 -\sum_{j\neq i}^d\frac{ \Eb[s_i^3]\eta_j }{2(|\alpha_{i}|+|\alpha_{j}|)}  (\b_j
	\b_i^\TR + \b_i\b_j^\TR),  \label{SYM3}\\
\R^{SYM}_{(4)}&=&\sum_{j\neq i}^d \frac{\beta_i - \gamma_i^2 + \beta_j - \gamma_j^2 + 
	\alpha_j^2 -\eta_i^2 - \eta_j^2
 	}{(|\alpha_i| + |\alpha_j|)^2}\b_j\b_j^\TR + \tau_i\b_i\b_i^\TR.  \label{SYM4}
\een
\end{theo}
\begin{coro}\label{coroSYM}For $i,j=1,\ldots,d$, there holds
$
N^{1/2}({\G}_{ij}^{SYM} - \delta_{ij})\CD \Nc(0, V_{(k)}^{SYM}),
$
 where 
\begin{enumerate}
\item Case $j=i$:
 \be
V^{SYM}_{(1)}=V^{SYM}_{(2)}=0 ,\quad\quad V^{SYM}_{(3)}=V^{SYM}_{(4)}=\tau_i. 
\ee
\item Case $j\neq i$:
\ben
V_{(1)}^{SYM}&=&\frac{\beta_i  + \beta_j - 2\gamma_i\gamma_j  }{(|\alpha_i| + |\alpha_j|)^2}, \quad\label{gain1} \\
V_{(2)}^{SYM}&=&\frac{\beta_i  + \beta_j - 2\gamma_i\gamma_j -2\eta_i^2  }{(|\alpha_i| + |\alpha_j|)^2}, \quad\label{gain2} \\
V_{(3)}^{SYM}&=&\frac{\beta_i - \gamma_i^2 + \beta_j - \gamma_j^2 + \alpha_j^2  }{(|\alpha_i| + |\alpha_j|)^2}, \quad\label{gain3} \\
V_{(4)}^{SYM}&=&\frac{\beta_i - \gamma_i^2 + \beta_j - \gamma_j^2 + \alpha_j^2 -\eta_i^2 - \eta_j^2
 }{(|\alpha_i| + |\alpha_j|)^2}. \quad\label{gain4}
\een
\end{enumerate}
\end{coro}
\begin{rema}
Although the asymptotic error of the FastICA algorithm has already been studied by quite a few researchers \cite{HYVA97,HYVA2006,OLLI2011,TICHOJA},
many of the results presented in this contribution, notably expressions (\ref{DFL1})-(\ref{DFL3}) established in Theorem \ref{main1} and (\ref{SYM1})-(\ref{SYM4}) 
in Theorem \ref{main2}, are new.
\end{rema}

\begin{figure}[t]
\centerline{
\includegraphics[width=1.3\textwidth]{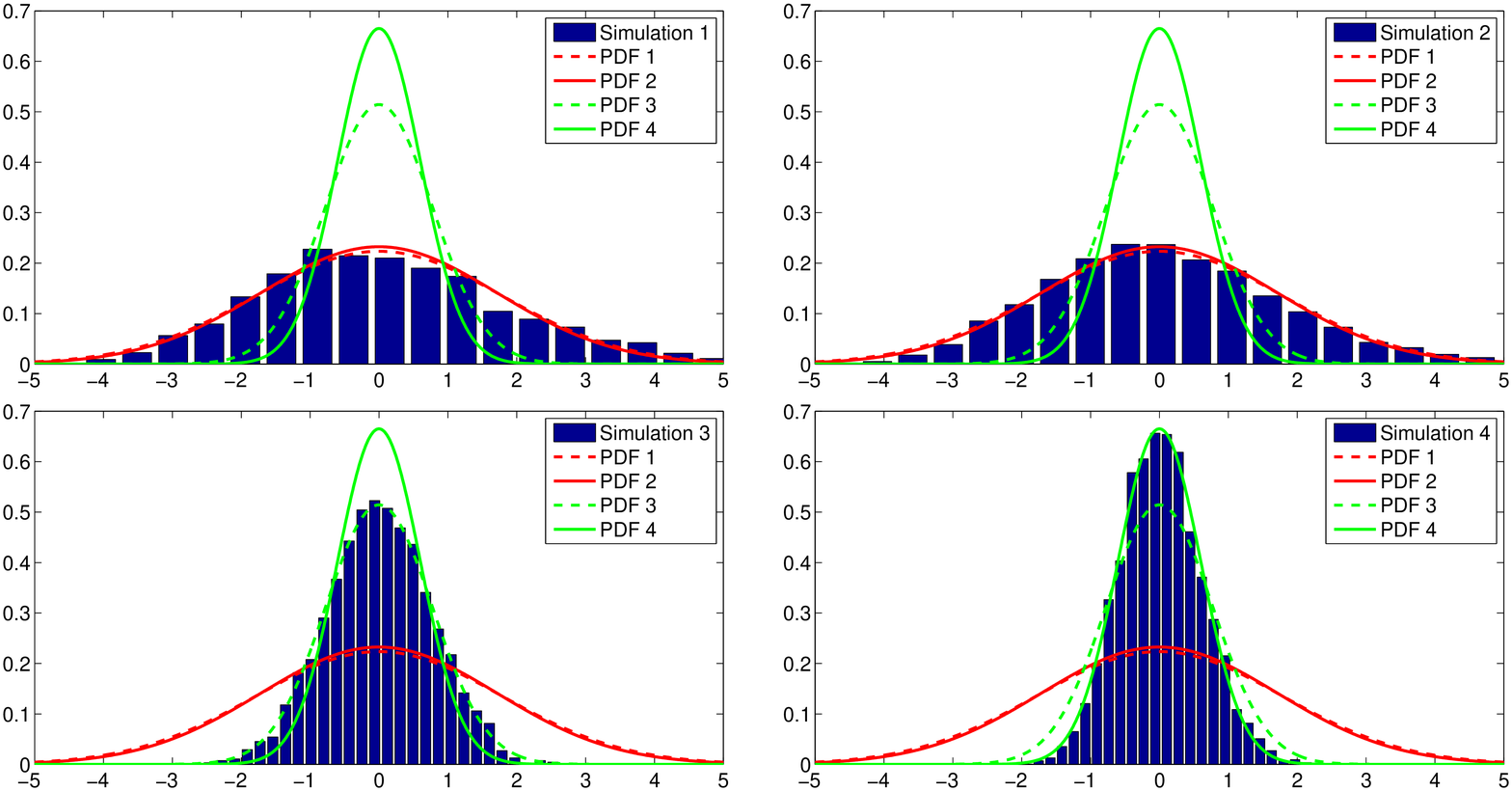}}
\caption{Asymptotic error of the deflationary FastICA in each preprocessing scenario.
We plotted the histograms of an (upper) off-diagonal entry of $N^{1/2}{\G}^{DFL}$
in 5000 independent trials versus
the theoretical curves of the Gaussian PDFs with variances given by
 (\ref{dflgain1})-(\ref{dflgain4}). \label{DFLplot}}
\centerline{
\includegraphics[width=1.3\textwidth]{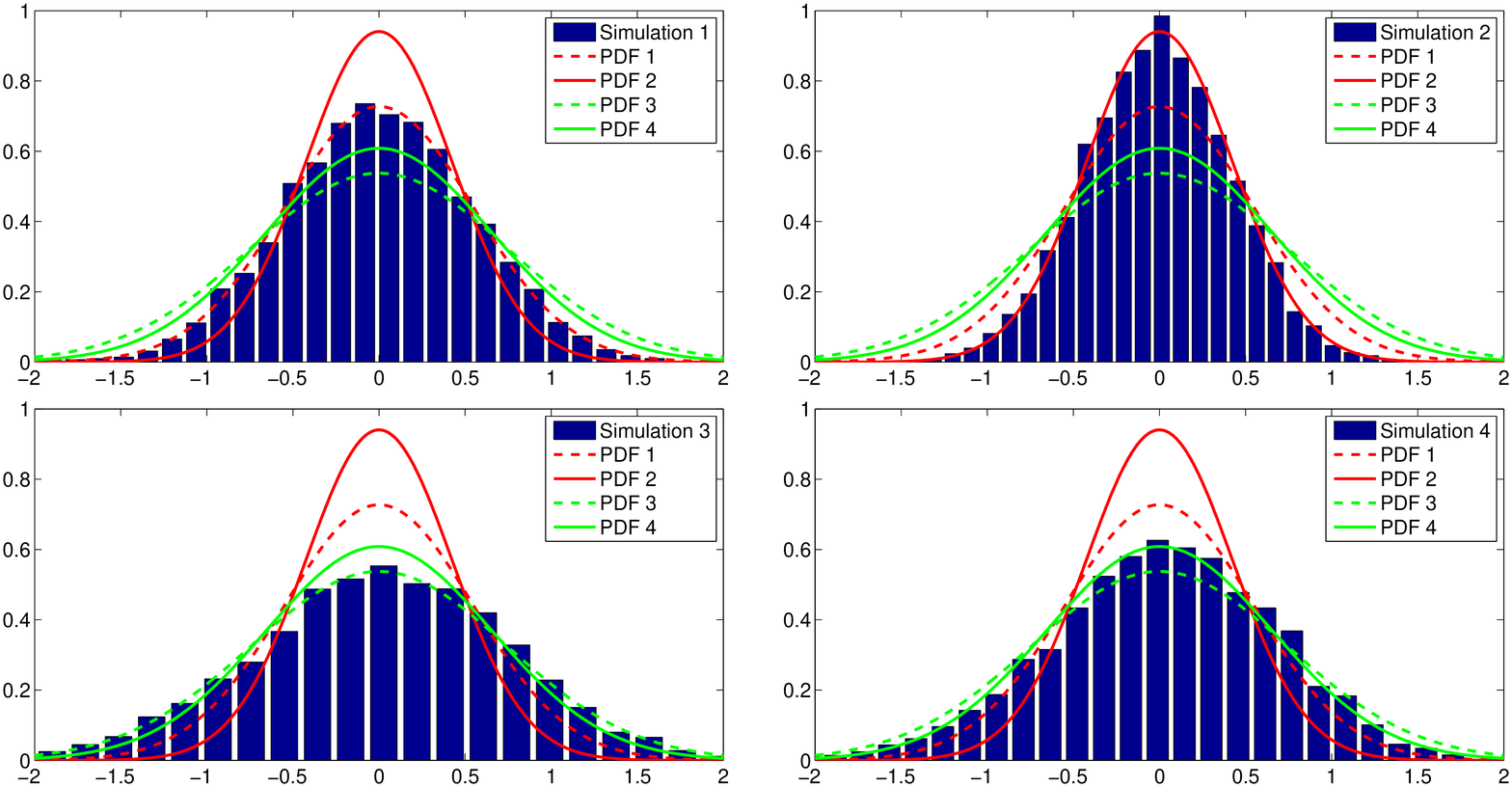}}
\caption{Asymptotic error of the symmetric FastICA in each preprocessing scenario.
We plotted the histograms of an off-diagonal entry of $N^{1/2}{\G}^{SYM}$
in 5000 independent trials versus
the theoretical curves of the Gaussian PDFs with variances given by
 (\ref{gain1})-(\ref{gain4}). \label{SYMplot}}
\end{figure}

\begin{exam}
The validity of formulas (\ref{dflgain1})-(\ref{dflgain4}) and (\ref{gain1})-(\ref{gain4}) is verified in computer simulations, see Fig. \ref{DFLplot} and Fig. \ref{SYMplot}. The simulations are configured as follows: $d=3$, $N=5000$, all three sources have identical bimodal Gaussian distribution with asymmetrical density. Both deflationary FastICA and symmetric FastICA have been tested with different data preprocessing (\ref{Preprocess1})-(\ref{Preprocess4}) in 5000 independent trials.
\end{exam}

\subsection{Discussion}
First, comparing the expressions
in Corollary \ref{coroDFL2} and Corollary \ref{coroSYM},
 we find that for the $(i,j)$th entry of the gain matrix,
\be
V^{DFL}_{(1)} - V^{DFL}_{(2)}&=& V^{DFL}_{(3)} - V^{DFL}_{(4)}
=\frac{\eta_j^2}{\alpha_j^2}, \quad j<i,\\
V^{DFL}_{(1)} - V^{DFL}_{(2)}&=& V^{DFL}_{(3)} - V^{DFL}_{(4)}
=\frac{\eta_i^2}{\alpha_i^2}, \quad j>i,\\
V^{SYM}_{(1)} - V^{SYM}_{(2)}&=& 
\frac{2\eta_i^2}{(|\alpha_i| + |\alpha_j|)^2},\quad i\neq j,\\
V^{SYM}_{(3)} - V^{SYM}_{(4)}&=& 
\frac{\eta_i^2 + \eta_j^2}{(|\alpha_i| + |\alpha_j|)^2},\quad i\neq j.
\ee
Since all the differences above are non-negative\footnote{They become zero if 
$\eta_i$ and/or $\eta_j$ vanish. This is the case
 if, e.g. $g$ is pair and the involved sources have symmetric distributions. 
 }, we assert that the empirical data centering generally leads to a better asymptotic performance.

\section{Conclusion}
The contribution of this work is twofold. First, we derived explicit formulas for the asymptotic error of the two most important variants of the FastICA algorithm, the deflationary FastICA and the symmetric FastICA, under four different data preprocessing scenarios. Many of the presented formulas are novel. 
Second, we assessed the impact of empirical data preprocessing procedure on the
asymptotic performance of the algorithms.
We showed that, compared to the theoretical data centering, the empirical data centering generally leads to a better separation performance.

\bibliographystyle{IEEEtran}
\bibliography{IEEEabrv,IeeeBib}

\begin{thebibliography}{1}
\providecommand{\url}[1]{#1}
\csname url@samestyle\endcsname
\providecommand{\newblock}{\relax}
\providecommand{\bibinfo}[2]{#2}
\providecommand{\BIBentrySTDinterwordspacing}{\spaceskip=0pt\relax}
\providecommand{\BIBentryALTinterwordstretchfactor}{4}
\providecommand{\BIBentryALTinterwordspacing}{\spaceskip=\fontdimen2\font plus
\BIBentryALTinterwordstretchfactor\fontdimen3\font minus
  \fontdimen4\font\relax}
\providecommand{\BIBforeignlanguage}[2]{{%
\expandafter\ifx\csname l@#1\endcsname\relax
\typeout{** WARNING: IEEEtran.bst: No hyphenation pattern has been}%
\typeout{** loaded for the language `#1'. Using the pattern for}%
\typeout{** the default language instead.}%
\else
\language=\csname l@#1\endcsname
\fi
#2}}
\providecommand{\BIBdecl}{\relax}
\BIBdecl

\bibitem{COMO94}
P.~Comon, ``Independent component analysis: a new concept?'' \emph{Signal
  Processing}, vol.~36, no.~3, pp. 287--314, Apr. 1994.

\bibitem{CARD1993}
J.~F. Cardoso and A.~Souloumiac, ``Blind beamforming for non-gaussian
  signals,'' \emph{IEEE Proceedings-F}, vol. 140, no.~6, pp. 362--370, Dec.
  1993.

\bibitem{HYVA99}
A.~Hyv{\"a}rinen, ``Fast and robust fixed-point algorithms for independent
  component analysis,'' \emph{IEEE Transactions on Neural Networks}, vol.~10,
  no.~3, pp. 626--634, 1999.

\bibitem{WEI4}
T.~Wei, ``A convergence and asymptotic analysis of the generalized symmetric
  fastica algorithm (submitted),'' \emph{ArXiv}, 2015.

\bibitem{HYVA97}
A.~Hyv{\"a}rinen, ``One-unit contrast functions for independent component
  analysis: A statistical analysis,'' in \emph{Proc. IEEE NNSP Workshop
  '97}.\hskip 1em plus 0.5em minus 0.4em\relax Neural Networks for Signal
  Processing VII, 1997.

\bibitem{HYVA2006}
A.~Shimizu, A.~Hyv{\"{a}}rinen, K.~Yutaka, P.~Hoyer, and A.~J. Kerminen,
  ``Testing signifcance of mixing and demixing coefficients in {ICA},'' in
  \emph{Int. Conf. Independent Component Analysis ({ICA} 2006)}, 2006.

\bibitem{OLLI2011}
K.~Nordhausen, P.~Ilmonen, A.~Mandal, H.~Oja, and E.~Ollila, ``Deflation-based
  {FastICA} reloaded,'' in \emph{19th European Signal Processing Conference
  (EUSIPCO 2011)}, Barcelona, Spain, Sep. 2011.

\bibitem{TICHOJA}
P.~Tichavsky, Z.~Koldovsky, and E.~Oja, ``Performance analysis of the {FastICA}
  algorithm and cramer-rao bounds for linear independent component analysis,''
  \emph{IEEE transactions on Signal Processing}, vol.~54, no.~4, pp.
  1189--1203, Apr. 2006.

\end{thebibliography}

\end{document}